\documentclass[letterpaper]{article} % DO NOT CHANGE THIS
\usepackage{aaai2026}  % DO NOT CHANGE THIS
\usepackage{times}  % DO NOT CHANGE THIS
\usepackage{booktabs}
\usepackage{amsmath,amsthm,amssymb}
\usepackage{tabularx}
\usepackage{enumitem}
\usepackage{placeins} 
\usepackage{helvet}  % DO NOT CHANGE THIS
\usepackage{courier}  % DO NOT CHANGE THIS
\usepackage[hyphens]{url}  % DO NOT CHANGE THIS
\usepackage{graphicx} % DO NOT CHANGE THIS
\usepackage{natbib}  % DO NOT CHANGE THIS AND DO NOT ADD ANY OPTIONS TO IT
\usepackage{caption} % DO NOT CHANGE THIS AND DO NOT ADD ANY OPTIONS TO IT
\usepackage{placeins}  
\usepackage{dblfloatfix}
\newcommand{\emails}[1]{#1}
\usepackage{algorithm}
\usepackage{algpseudocode} 
\usepackage{breqn}
\theoremstyle{definition}
\newtheorem{definition}{Definition}
\theoremstyle{plain}
\newtheorem{theorem}{Theorem}
\newtheorem{lemma}{Lemma}

\theoremstyle{remark}

\usepackage{caption}
\usepackage{makecell}

\usepackage{newfloat}
\usepackage{listings}
\DeclareCaptionStyle{ruled}{labelfont=normalfont,labelsep=colon,strut=off} % DO NOT CHANGE THIS
\floatstyle{ruled}
\newfloat{listing}{tb}{lst}{}
\floatname{listing}{Listing}

\title{Learning-Augmented Ski Rental with Discrete Distributions: A Bayesian Approach}

\author{
    Bosun Kang\textsuperscript{\rm 1,2}\;
    Hyejun Park\textsuperscript{\rm 1,3}\;
    Chenglin Fan\textsuperscript{\rm 1,3}\thanks{Corresponding author.}
}

\affiliations{
    \textsuperscript{\rm 1}Algorithmic Foundations of Data Science Lab\\
    \textsuperscript{\rm 2}Department of Liberal Studies,\\
    Seoul National University, South Korea\\
    \textsuperscript{\rm 3}Department of Computer Science and Engineering,\\
    Seoul National University, South Korea\\[0.5ex]
    \emails{\{bosun1043, 1253spring, fanchenglin\}@snu.ac.kr}
}

\begin{document}

\maketitle

\begin{abstract}

We revisit the classic ski rental problem through the lens of Bayesian decision-making and machine-learned predictions. While traditional algorithms minimize worst-case cost without assumptions, and recent learning-augmented approaches leverage noisy forecasts with robustness guarantees, our work unifies these perspectives. We propose a discrete Bayesian framework that maintains exact posterior distributions over the time horizon, enabling principled uncertainty quantification and seamless incorporation of expert priors. Our algorithm achieves prior-dependent competitive guarantees and gracefully interpolates between worst-case and fully-informed settings. 
Our extensive experimental evaluation demonstrates superior empirical performance across diverse scenarios, achieving near-optimal results under accurate priors while maintaining robust worst-case guarantees. This framework naturally extends to incorporate multiple predictions, non-uniform priors, and contextual information, highlighting the practical advantages of Bayesian reasoning in online decision problems with imperfect predictions.

\end{abstract}

\section{Introduction}

The ski rental problem is a foundational challenge in online algorithms, capturing the fundamental trade-off between short-term flexibility and long-term commitment under uncertainty. A decision-maker must repeatedly choose between renting an item at a per-use cost or purchasing it upfront at a fixed cost, without knowing how long the item will be needed. Originally introduced by~\citet{karlin1994competitive}, this problem models a wide range of real-world scenarios, including cloud resource provisioning, equipment leasing, and inventory management.

In the classical setting, no deterministic algorithm can achieve a competitive ratio better than $2$, and randomization improves this to $\frac{e}{e-1} \approx 1.582$~\citep{chrobak1991new}. However, these guarantees are worst-case and fail to leverage any structure or prior information about the underlying demand. Recent work on \emph{learning-augmented algorithms} has proposed integrating machine-learned predictions to improve online performance while retaining robustness~\citep{LykourisV18, mitzenmacher2020consistency, purohit2018improving}. Yet existing approaches often rely on point estimates, regime heuristics, or hand-crafted thresholds, which are brittle in the face of prediction noise and difficult to analyze rigorously.

This paper introduces a Bayesian framework for the ski rental problem that is both learning-augmented and fully probabilistic. Our approach maintains a posterior distribution over the unknown time horizon and updates it exactly each day based on observed survival. This enables decisions based on expected utility rather than point forecasts, and provides a principled mechanism for incorporating prior knowledge—be it empirical, structured, or even adversarially incorrect. Throughout the paper we assume the unknown horizon is bounded by a known $M$.

\subsection{Contributions}
This work makes several key contributions.
First, we introduce the first learning-augmented ski rental algorithm based on exact discrete Bayesian inference, removing the need for point estimates or regime switching.
Second, we establish theoretical guarantees on competitive ratio and regret under classical priors, showing robustness to prior misspecification.
Third, extensive experiments on synthetic and real-world priors confirm consistent outperformance over deterministic, randomized, and prediction-based baselines—even under noisy or adversarial conditions. Our Bayesian formulation is modular and extensible, supporting multiple priors, structured uncertainty, and continuous-time variants, while adapting to multi-modal priors via decisions informed by mode weights and positions.

\subsection{Technical Overview}

\paragraph{Notation.}
Consider an unknown season length $T \in \{1,\dots,M\}$ with prior mass function
$\pi_k \triangleq \Pr[T = k]$ for $k=1,\dots,M$.
Using $k$ to denote a realized trip length (horizon).
Let $b>1$ be the one-time \emph{buy cost} and, unless stated otherwise,
the per-day \emph{rental cost} is normalized to $1$.

It maintains the posterior over $T$ conditioned on survival ($T\ge t$) and compares the expected remaining rental cost with $b$.

\begin{enumerate}
  \item \textbf{Posterior update:} $\Pr(T = k \mid T \ge t) = \dfrac{\pi_k}{\sum_{j = t}^{M} \pi_j}$ for $k \ge t$.
    \item \textbf{Expected rental cost:} 
   $E_{\text{rent}}(t) = \sum_{k = t}^{M} \Pr(T = k \mid T \ge t) \cdot (k - t + 1)$.
  \item \textbf{Decision:} If $b \le E_{\text{rent}}(t)$, buy; else, rent and proceed to day $t+1$.
\end{enumerate}

\paragraph{Units.}
With per-day rental cost normalized to $1$, $b$ and $k-t+1$ are directly comparable:
\begin{equation}
%E_{\mathrm{rent}}(t)
%= \sum_{k=t}^{M} \Pr[T=k \mid T\ge t]\cdot \min\{\,k-t+1,\; b\,\}.
E_{\mathrm{rent}}(t)
= \sum_{k=t}^{M} \Pr[T=k \mid T\ge t]\cdot (k-t+1).
\label{eq:Erent}
\end{equation}

This fully Bayesian strategy naturally quantifies uncertainty, makes decisions by expected-utility maximization, and seamlessly integrates any prior information. In the sections that follow, we formalize the algorithm, prove its competitive guarantees, and explore its empirical performance and extensions.

We begin by reviewing the classical ski rental problem, which forms the foundation for our Bayesian extension.

\subsection{Related Work}

\textbf{Classical Results.} 
The ski rental problem was formalized by \citet{karlin1994competitive}, who showed a competitive ratio lower bound of 2 for deterministic algorithms, achievable via a simple threshold strategy. \citet{chrobak1991new} proved that randomization improves this to $e/(e-1) \approx 1.582$. Extensions include multi-item \citep{fleischer2001optimal} and dynamic-cost settings \citep{young2000online}.

\textbf{Learning-Augmented Algorithms.} 
Recent work integrates predictions into online algorithms, with frameworks balancing consistency and robustness \citep{LykourisV18,mitzenmacher2020consistency}. For ski rental, \citet{improvingOnlineML} proposed prediction-dependent thresholds, while \citet{anand2022online} studied multiple predictions. More recent advances extend to offline problems such as matching~\cite{dinitz2021faster}, clustering~\cite{ergun2021learning}, and sorting~\cite{bai2023sorting}. Inspired by learned index structures~\cite{kraska2018case}, robust techniques have emerged for combining multiple predictions~\cite{anand2022online, antoniadis2023mixing}, alongside new applications in binary search~\cite{dinitz2024binary}, warm-started decision-making~\cite{blum2025competitive}, and learning-informed dynamic graphs~\cite{brand2024dynamic}.

\textbf{Bayesian Approaches.} 
Bayesian methods are well-established in offline problems (e.g., Thompson sampling \citep{agrawal2013thompson}), but competitive analyses for online settings remain rare.

\textbf{Our Work.} 
To the best of our knowledge, our work is the first to introduce a fully Bayesian framework for the ski rental problem with distributional predictions, establishing provable guarantees under a discrete prior.

\section{Preliminaries}

We briefly review the classical ski rental problem and then present our discrete Bayesian extension.

\subsection{Classical Ski Rental}

At each day $t = 1, 2, \dots$, a skier chooses to:
\[
\begin{array}{ll}
\text{Rent:} & \text{pay \$1 for the day},\\
\text{Buy:} & \text{pay a one-time cost } b > 1 \text{ for unlimited use}.
\end{array}
\]
The offline optimum for an unknown season length $T$ is $\mathrm{OPT}(T) = \min\{T, b\}$. An online algorithm $\mathcal{A}$ incurs cost $\mathrm{ALG}(\mathcal{A}, T)$, and its competitive ratio is
\[
\mathrm{CR}(\mathcal{A}) = \sup_{T \ge 1} \frac{\mathrm{ALG}(\mathcal{A}, T)}{\mathrm{OPT}(T)}.
\]
No deterministic algorithm achieves $\mathrm{CR} < 2$, but randomization attains the optimal $e/(e - 1) \approx 1.582$.

\subsection{Discrete Bayesian Extension}

We assume a finite horizon $M$ and prior $\Pr[T = k] = \pi_k$ for $k = 1, \dots, M$, with $\sum_{k = 1}^M \pi_k = 1$.  
Conditioned on survival ($T \ge t$), the posterior becomes
\[
\Pr[T = k \mid T \ge t] = \frac{\pi_k}{\sum_{j = t}^M \pi_j}, \quad k = t, \dots, M.
\]
Define the expected remaining rental cost:
\[
E_{\mathrm{rent}}(t) = \sum_{k = t}^M \Pr[T = k \mid T \ge t] \cdot (k - t + 1).
\]
Our policy purchases on the first day $t$ such that
$b \le E_{\mathrm{rent}}(t),
$
and continues renting otherwise, reassessing daily.

\subsection{Computational Complexity}

This section provides a detailed analysis of the computational complexity of the proposed Bayesian decision procedure, including its time, space, and sparse-support implementation aspects.

\begin{enumerate}
    \item[1)] \textbf{Time:} Computing the normalization factor and the expected-cost sum each day takes $O(M)$ time, resulting in a worst-case total time complexity of $O(M^2)$.

    \item[2)] \textbf{Space:} Storing the prior and posterior distributions requires $O(M)$ memory. All arithmetic operations involve nonnegative sums and divisions, ensuring numerical stability.

    \item[3)] \textbf{Sparse-Support Implementation:} When the prior has $n$ (with $n \le M$) nonzero entries $(k_i, \pi_{k_i})$, we can optimize the algorithm by indexing the support points using a balanced BST or hash map, maintaining prefix sums over survival probabilities for $O(\log n)$ query time, and computing $E_{\mathrm{rent}}(t)$ by iterating only over the $n$ support points. These optimizations reduce the overall runtime to $O(n \log n)$ without affecting the competitive guarantees.
\end{enumerate}

With these preliminaries established, we now present the detailed Bayesian decision procedure that leverages this probabilistic framework to make optimal online decisions.

\section{Bayesian Decision Procedure}

\subsection{Algorithm Description}
Our approach makes decisions via exact Bayesian inference over a discrete prior on the unknown horizon. As shown in Algorithm~\ref{alg:bayesian_detailed}, given a buy cost \(b > 1\), horizon bound \(M\), and prior \(\pi = [\pi_1, \ldots, \pi_M]\), the algorithm updates its posterior over \([t, M]\) at each day \(t\), conditioned on survival.
It computes the expected rental cost under this posterior and compares it to \(b\). If the expected cost exceeds \(b\), it buys; otherwise, it rents and continues. This repeats until purchase or horizon end.
This simple, efficient method adapts to arbitrary discrete priors without relying on heuristics or point estimates, and naturally extends to contextual or time-varying settings.
% ==== Display equations referenced inside the algorithm ====
\begin{subequations}\label{eq:bayes_update}
\begin{align}
Z_t &\triangleq \sum_{k=t}^{M} \pi_k, \label{eq:bayes_update_Zt}\\
p_{t,k} &\triangleq \frac{\pi_k}{Z_t} \quad (k=t,\dots,M), \label{eq:bayes_update_ptk}\\
E_{\mathrm{rent}}(t) &\triangleq \sum_{k=t}^{M} p_{t,k} \times(k-t+1). \label{eq:bayes_update_E}
\end{align}
\end{subequations}

\begin{algorithm}[H]
\caption{Discrete Bayesian Ski Rental}
\label{alg:bayesian_detailed}
\begin{algorithmic}[1]
\Require Buy cost $b>1$, horizon bound $M\ge 1$, prior distribution $\pi=[\pi_1,\pi_2,\ldots,\pi_M]$
\Ensure Purchase day $t^*$ (or $M{+}1$ if no purchase occurs)

\State $t \leftarrow 1$
\State \textbf{Validate/normalize prior:} \If{$\sum_{k=1}^{M}\pi_k = 0$} \State \Return $M{+}1$ \Comment{Invalid prior} \EndIf
\State $\pi \leftarrow \pi / \sum_{k=1}^{M}\pi_k$ \Comment{Normalize to a valid pmf}

\While{$t \le M$}
    \State Compute $Z_t$, $p_{t,k}$, and $E_{\mathrm{rent}}(t)$ as in \eqref{eq:bayes_update}
    \If{$Z_t = 0$}
        \State \Return $M{+}1$ \Comment{No remaining probability mass}
    \EndIf
    \If{$b \le E_{\mathrm{rent}}(t)$}
        \State \Return $t$ \Comment{Buy today}
    \Else
        \State $t \leftarrow t+1$ \Comment{Rent today; re-evaluate tomorrow}
    \EndIf
\EndWhile

\State \Return $M{+}1$ \Comment{Never buy within horizon}
\end{algorithmic}
\end{algorithm}

\subsection{Algorithm Properties}

The algorithm operates by making a daily decision: whether to buy skis on day $t$ or continue renting. On each day $t$, it compares the known cost of buying, which is $b$, against the expected cost of continuing to rent from day $t$ onward, denoted by $E_{\text{rent}}(t)$. This expected rental cost is computed with respect to the posterior distribution over ski days, conditional on the fact that skiing has lasted at least until day $t$. The decision rule is simple: buy as soon as $b \leq E_{\text{rent}}(t)$. This policy provably minimizes the expected total cost under $\pi$.
\begin{lemma}[Monotonicity of Purchase Incentive]
Under the log-concave prior distribution, the expected remaining rental cost $E_{\text{rent}}(t)$ is non-increasing in $t$:
\[
E_{\text{rent}}(t+1) \leq E_{\text{rent}}(t).
\]
\end{lemma}

We analyze the performance of this algorithm in terms of its \emph{expected competitive ratio} (ECR), defined as the ratio between the algorithm’s expected cost and the expected cost of the offline optimal strategy:
\begin{align}
\text{ECR}
&= \frac{
    \sum_{k=1}^{t^*-1} \pi_k \, k
    + \sum_{k=t^*}^{M} \pi_k \, (t^*-1 + b)
}{%
    \sum_{k=1}^{M} \pi_k \, \min(k, b)
}.
\end{align}

The offline optimal algorithm, knowing $k$ in advance, incurs cost $\min(k, b)$. Its expected cost is therefore
\[
\mathbb{E}_{\pi}[\text{Cost}_{\text{OPT}}] = \sum_{k=1}^M \pi_k \cdot \min(k, b).
\]

Let $t^*$ denote the day on which the Bayes-optimal algorithm chooses to buy. Its expected cost is
\[
\mathbb{E}_{\pi}[\text{Cost}_{\text{ALG}}] = \sum_{k=1}^{t^*-1} \pi_k \cdot k + \sum_{k=t^*}^{M} \pi_k \cdot (t^*-1 + b).
\]

Thus, the expected competitive ratio becomes:
\[
\text{ECR} = \frac{\sum_{k=1}^{t^*-1} \pi_k \cdot k + \sum_{k=t^*}^M \pi_k \cdot (t^*-1 + b)}{\sum_{k=1}^{M} \pi_k \cdot \min(k, b)}.
\]

This ratio is always at least $1$, and equals $1$ only when $\pi$ is concentrated on a single value of $k$ such that the algorithm makes the correct offline choice.

Compared to classical algorithms, this Bayesian approach provides significantly improved performance in expectation. The deterministic strategy of buying on day $b$ has a worst-case competitive ratio of $(2b - 1)/b$, which approaches $2$ as $b \to \infty$. The optimal randomized strategy achieves a worst-case ratio of $e/(e-1) \approx 1.58$. In contrast, the Bayes-optimal strategy leverages prior information to minimize expected cost and achieves strictly better performance when the prior is informative. However, this benefit is contingent on the quality of the prior: a poorly specified prior may result in suboptimal outcomes.

\section{ When Bayesian Outperforms Classical Algorithms}

Having established the core algorithm and its properties, we now analyze specific scenarios where the Bayesian approach significantly outperforms classical algorithms, demonstrating its practical advantages under various prior distributions. 

\subsection*{Case 1: Uniform Prior (Fixed Range)}

We analyze the scenario where the trip length $k$ is uniformly distributed over a fixed, finite range, $k \in \{1, 2, \dots, N\}$:
\[
\pi_k = \frac{1}{N}, \quad \text{for } k \in \{1, 2, \dots, N\}.
\]

\noindent\textbf{Bayesian Algorithm's Behavior.}  
The Bayesian algorithm makes its decision entirely on \emph{Day 1}. Since the expected remaining rental cost $E_{\text{rent}}(t)$ is non-increasing in $t$ (as shown by the Monotonicity Lemma), if the algorithm does not buy on Day~1, it will \emph{never} buy.  

The expected rental cost on Day~1 is given by the expectation of the uniform prior:
\[
E_{\mathrm{rent}}(1) = \sum_{k=1}^{N} \pi_k \cdot k 
= \frac{1}{N} \sum_{k=1}^{N} k  
= \frac{N+1}{2}.
\]
Therefore, the algorithm chooses to buy on Day~1  ($t^* = 1$) if and only if
\[
b \le E_{\mathrm{rent}}(1),
\]
and otherwise never buys ($t^* = N + 1$) when
\[
b > E_{\mathrm{rent}}(1).
\]
The following theorem holds.

\begin{theorem}    
[Uniform Prior]
If \(\pi_k = 1/N\) for \(k \in \{1, \ldots, N\}\), the Bayesian algorithm achieves an ECR of  \[
\text{ECR} =
\begin{cases}
1, & \text{if } N \leq b, \\[6pt]
\dfrac{N(N + 1)}{b(2N - b + 1)}, & \text{if } b < N < 2b - 1, \\[8pt]
\dfrac{2N}{2N - b + 1}, & \text{if } N \geq 2b - 1.
\end{cases}
\].
\end{theorem}

\subsection{Case 2: Geometric Prior with Fixed Support}

%We now analyze the performance of the Bayesian algorithm under a fixed-range geometric prior.
This setting assumes the trip length $k$ is drawn from a geometric distribution truncated to the range $k \in \{1, 2, \dots, N\}$. 
%Crucially, this prior exhibits an \emph{Increasing Hazard Rate (IHR)}, which implies that the expected remaining rental cost $E_{\mathrm{rent}}(t)$ is non-increasing. Consequently, the optimal decision simplifies to a single comparison on Day~1.

\subsubsection*{Prior Distribution and Expected Costs}

The prior distribution over trip lengths is defined as:
\[
\pi_k = \frac{p(1-p)^{k-1}}{1 - (1-p)^N}, \quad \text{for } k = 1, 2, \dots, N,
\]
where $p$ denotes the success probability of the geometric distribution.

\begin{align*}
\text{ECR}
&= \frac{
    \sum_{\tau=1}^N \pi_\tau \left[
        \min\!\bigl(t^*(\tau), \tau\bigr) - 1
        + b \cdot \mathbf{1}_{\{t^*(\tau) \leq \tau\}}
    \right]
}{%
    \sum_{\tau=1}^N \pi_\tau \, \min(\tau, b)
}.
\end{align*}
where \(t^*(\tau)\) is the stopping time when the true horizon is \(\tau\), determined by:
\[
t^*(\tau) = \min\left\{ t \leq \tau : b \leq \sum_{k=t}^N \frac{\pi_k}{Z_t} (k-t+1) \right\}.
\]

Let $E_1$ denote the expected trip length (the mean of the truncated geometric distribution):
\[
E_1 \triangleq E_{\mathrm{rent}}(1)
= \sum_{k=1}^N k \pi_k.
\]
We have 
\[
E_{\mathrm{rent}}(k+1) = \frac{1}{p} - \frac{(N-k) (1-p)^{N-k}}{1 - (1-p)^{N-k}}
\]
is  decreasing in  $k$ when  
$N$ is larger than some constant.

\begin{theorem}[Truncated Geometric Prior]
If the prior is given by $\pi_k \propto p(1-p)^{k-1}$ truncated to $k \le N>2$, the Bayesian algorithm achieves an Expected Competitive Ratio 
$ECR=\dfrac{\min(E_1,b)}{\sum_{k=1}^N \pi_k \min(k,b)}.$

\end{theorem}

\subsection{Case 3: Truncated Gaussian Prior}

We assume $T$ is drawn from a truncated Gaussian prior on $\{1,\dots,N\}$:

\paragraph{Prior Distribution.}  
Let $\mu$ and $\sigma$ be the mean and standard deviation of the untruncated Gaussian. The truncated Gaussian prior is defined over $\{1, 2, \dots, N\}$ as:
\[
\pi_k = \frac{e^{-\frac{(k-\mu)^2}{2\sigma^2}}}{\sum_{j=1}^N e^{-\frac{(j-\mu)^2}{2\sigma^2}}}, \quad \text{for } k \in \{1, 2, \dots, N\}.
\]
This prior is highly informative and places most of its probability mass near the mean $\mu$.
A discretized truncated Gaussian prior 
$
\pi_k \propto \exp\!\left(-\frac{(k-\mu)^2}{2\sigma^2}\right)
$
is log-concave; a log-concave discrete PMF has an increasing discrete hazard rate
$
h(t) = \frac{\pi_t}{\sum_{k=t}^{M} \pi_k},
$
and an increasing hazard rate (IHR) implies that the mean residual life  is nonincreasing.
Hence the Bayesian algorithm makes its decision entirely on \emph{Day 1}.

 We now extend the basic framework to handle more complex scenarios, including multiple predictions, adaptive learning, and contextual information.

\section{Algorithmic Extensions and Variants}

\subsection{Multiple Predictions Algorithm}

%When multiple independent predictions $\hat{T}_1, \hat{T}_2, \ldots, \hat{T}_n$ are available, we can optimally combine them as follows:

Given multiple independent predictions $\hat{T}_1, \ldots, \hat{T}_n$ about the unknown horizon $T$, each with an associated uncertainty level $\sigma_i$, we seek to compute a refined posterior distribution over $T$ that integrates all available information. We assume each prediction is a noisy observation of the true horizon, modeled as a Gaussian centered at $T$ with variance $\sigma_i^2$. Starting from a uniform prior, we sequentially update the posterior using Bayes' rule. Each prediction contributes a likelihood term, and the updates ensure that the final posterior distribution reflects all predictions in a statistically optimal way. The resulting posterior can then be used to make decisions in the ski rental problem with improved accuracy.

\begin{algorithm}[t]
\caption{Bayesian Ski Rental with Multiple Predictions}
\label{alg:multiple_predictions}
\begin{algorithmic}[1]
\Require Predictions $\{\hat{T}_1, \ldots, \hat{T}_n\}$ and their accuracies $\{\sigma_1, \ldots, \sigma_n\}$
\Ensure Combined posterior distribution

\State \textbf{Initialize:} Uniform prior $\pi_k^{(0)} = 1/M$
\For{$i = 1$ to $n$}
    \State \textbf{Compute Likelihood:} $L_i(k) = \frac{1}{\sqrt{2\pi\sigma_i^2}} \exp\left(-\frac{(k - \hat{T}_i)^2}{2\sigma_i^2}\right)$
    \State \textbf{Update Posterior:} $\pi_k^{(i)} \propto \pi_k^{(i-1)} \cdot L_i(k)$
    \State \textbf{Normalize:} $\pi_k^{(i)} = \pi_k^{(i)} / \sum_{j=1}^M \pi_j^{(i)}$
\EndFor
\State \textbf{return} $\pi^{(n)}$
\end{algorithmic}
\end{algorithm}
Our approach above not only leverages each predictor’s uncertainty but also achieves a near-optimal decision quality in expectation.

\subsection{Adaptive Prior Learning}

In many real-world scenarios, the ski rental problem is encountered repeatedly with varying horizons drawn from an unknown distribution. 
We propose an adaptive algorithm that learns this prior online, starting from a uniform initialization and updating it each round using the observed buying time $T_r$ via an exponential moving average with learning rate $\alpha$. 
This refinement enables adaptation to the underlying horizon distribution and improved decision-making, aligning with sequential or learning-augmented online optimization frameworks~\cite{gummadi2021sequential,lattanzi2020online}.

\begin{algorithm}[H]
\caption{Adaptive Bayesian Ski Rental}
\label{alg:adaptive}
\begin{algorithmic}[1]
\Require Number of rounds $R$, learning rate $\alpha$
\Ensure Sequence of purchasing decisions

\State \textbf{Initialize:} $\pi^{(1)} = \text{Uniform}(1, M)$
\For{round $r = 1$ to $R$}
    \State Run Bayesian algorithm using prior $\pi^{(r)}$
    \State Observe realized horizon $T_r$
    \State \textbf{Update:} $\pi_k^{(r+1)} = (1 - \alpha_r)\pi_k^{(r)} + \alpha_r \cdot \mathbf{1}[k = T_r]$
\EndFor
\end{algorithmic}
\end{algorithm}

\begin{theorem}[Adaptive Regret Bound]
With learning rate \[
\alpha_r = \min\left\{\,1,\ \sqrt{\frac{\log M}{r}}\,\right\},
\]
the adaptive algorithm incurs regret:
\[
\text{Regret}_R = \sum_{r=1}^R \left[\text{Cost}_r - \text{OPT}_r\right] = O(\sqrt{R \log M}).
\]
\end{theorem}

The theorem shows that the adaptive algorithm achieves cumulative regret 
\(O(\sqrt{R \log M})\) over \(R\) rounds, where \(\text{Cost}_r\) is the incurred cost 
and \(\text{OPT}_r\) is the offline optimal cost for horizon \(T_r\).

When contextual information $x \in \mathcal{X}$ is available, the prior can be 
conditioned on $x$ through a softmax parameterization:

\begin{definition}[Contextual Prior]
Given context $x$, the prior is defined as
\[
\pi_k(x) 
= \frac{\exp(\theta_k^\top \phi(x))}
       {\sum_{j=1}^M \exp(\theta_j^\top \phi(x))},
\]
where $\phi(x)$ is a feature map and $\theta_k$ is the parameter associated with 
horizon $k$.
\end{definition}

This parametrization lets the prior depend smoothly on $x$, effectively modeling 
$\pi(\cdot \mid x)$ as a categorical exponential-family distribution.
The expected remaining rental cost then becomes
\[
\mathbb{E}[T - t + 1 \mid x,\, T \ge t],
\]
so contextual information directly shifts the decision boundary through the 
induced posterior.

\section{Experiments}

We structure our experimental evaluation around the following questions:

\noindent\textbf{Q1. Robustness to prior misspecification.}  
How stable is the Bayesian algorithm when the assumed prior differs from the true horizon distribution?

\noindent\textbf{Q2. Performance under perfect prior knowledge.}  
How close does the algorithm get to the offline optimal when the prior is accurate?

\noindent\textbf{Q3. Noisy single predictions.}  
Can the algorithm maintain performance when only a biased or noisy point prediction is available?

\noindent\textbf{Q4. Multi-modal prior distributions.}  
Does the algorithm adapt to complex, multi-peaked distributions without explicit mode detection?

\vspace{0.5em}
\subsection{Experimental Setup}

We evaluate all methods with buy cost $b = 100$, horizon bound $M = 500$, 
and 10,000 Monte Carlo trials per configuration.  
We compare the Bayesian algorithm against deterministic thresholding, the optimal randomized strategy, 
point-prediction purchase, and the learning-augmented strategy of Kumar et al.~(2024).  
Performance is measured via competitive ratio (CR) and success rate.

\vspace{0.5em}
\subsection{Q1. Robustness to Prior Misspecification}

We test three Gaussian uncertainty regimes ($\sigma/\mu \in \{0.42, 0.33, 0.31\}$) and introduce 
misspecification by perturbing the mean, variance, and distributional shape.  
As shown in Figure~\ref{fig:prior_misspec}, performance degrades only mildly: 
the average cost increase is 5.3\%, and the worst-case degradation reaches 18.7\% under extreme mean errors.  
Variance and model-form errors have negligible impact.

\vspace{0.5em}
\subsection{Q2. Performance Under Perfect Prior Knowledge}

Table~\ref{tab:perfect_prior} summarizes performance when the assumed prior matches the true distribution.  
The Bayesian method achieves near-optimal CR $\approx 1.02$ with a 98.7\% success rate, 
significantly outperforming classical baselines.

\vspace{0.5em}
\subsection{Q3. Noisy Single Predictions}

We next evaluate robustness when only a single noisy prediction $\hat{T}$ is provided.
Predictions follow $\hat{T} \sim \mathcal{N}(\alpha T, (\beta T)^2)$ with $\beta = 0.3$ and bias $\alpha \in [0.5, 2.0]$.  
Table~\ref{tab:noisy_single_prediction} shows that the Bayesian method degrades smoothly 
from CR 1.05 to 1.43 as bias increases, consistently outperforming the point-prediction baseline.

\vspace{0.5em}
\subsection{Q4. Multi-modal Prior Distributions}

\begin{figure}[ht]
  \centering
  \caption{Comparison of multi-modal priors (left) and their survival functions (right).  
  Despite strong multi-modality in the density, the Bayesian threshold depends only on the 
  integrated survival mass $S(t)$, producing stable and coherent purchase decisions.}
  \vspace{0.6em}

  % Left figure
  \begin{minipage}{0.48\linewidth}
    \centering
    \includegraphics[width=\linewidth]{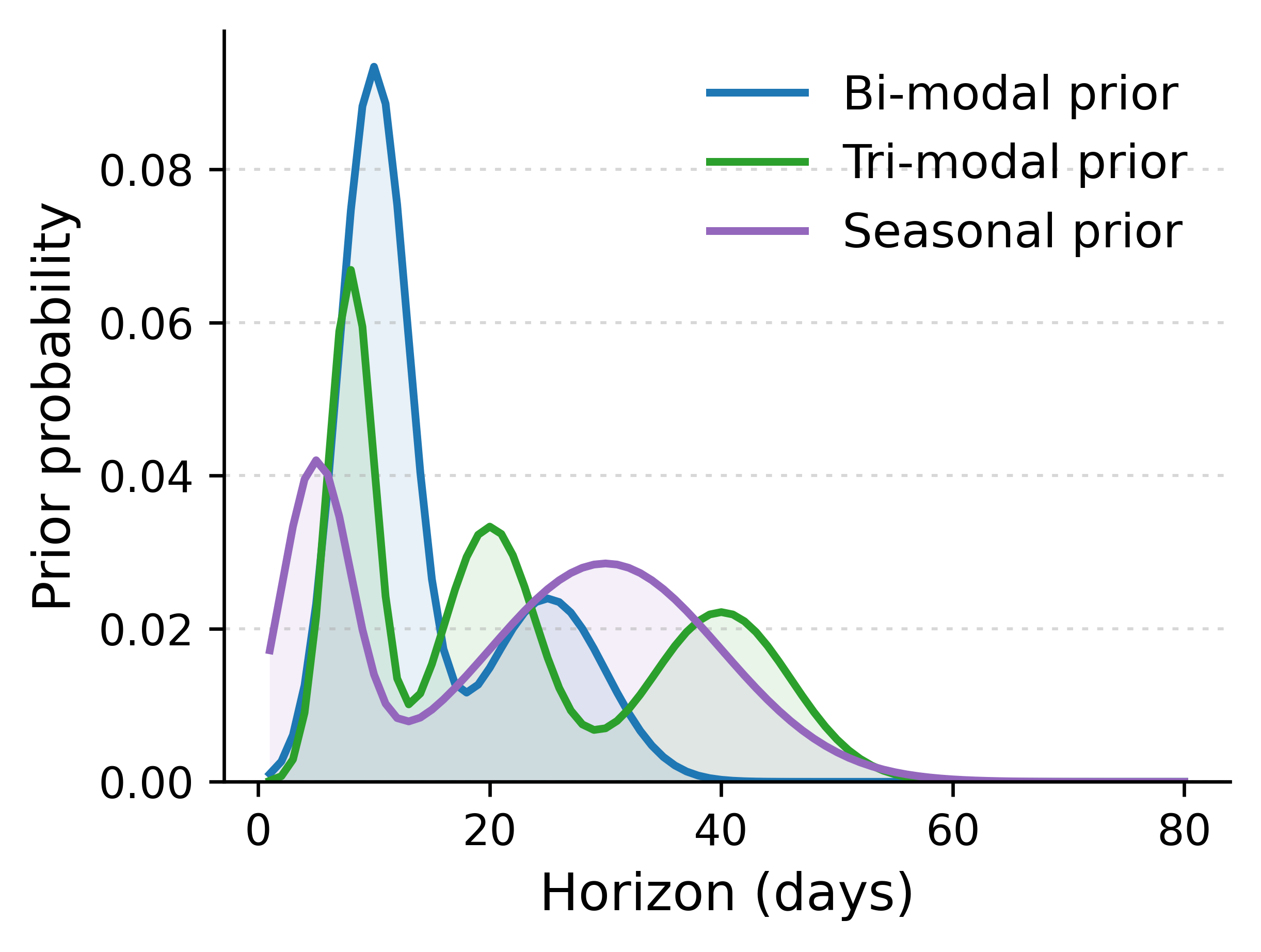}
  \end{minipage}
  \hfill
  % Right figure
  \begin{minipage}{0.48\linewidth}
    \centering
    \includegraphics[width=\linewidth]{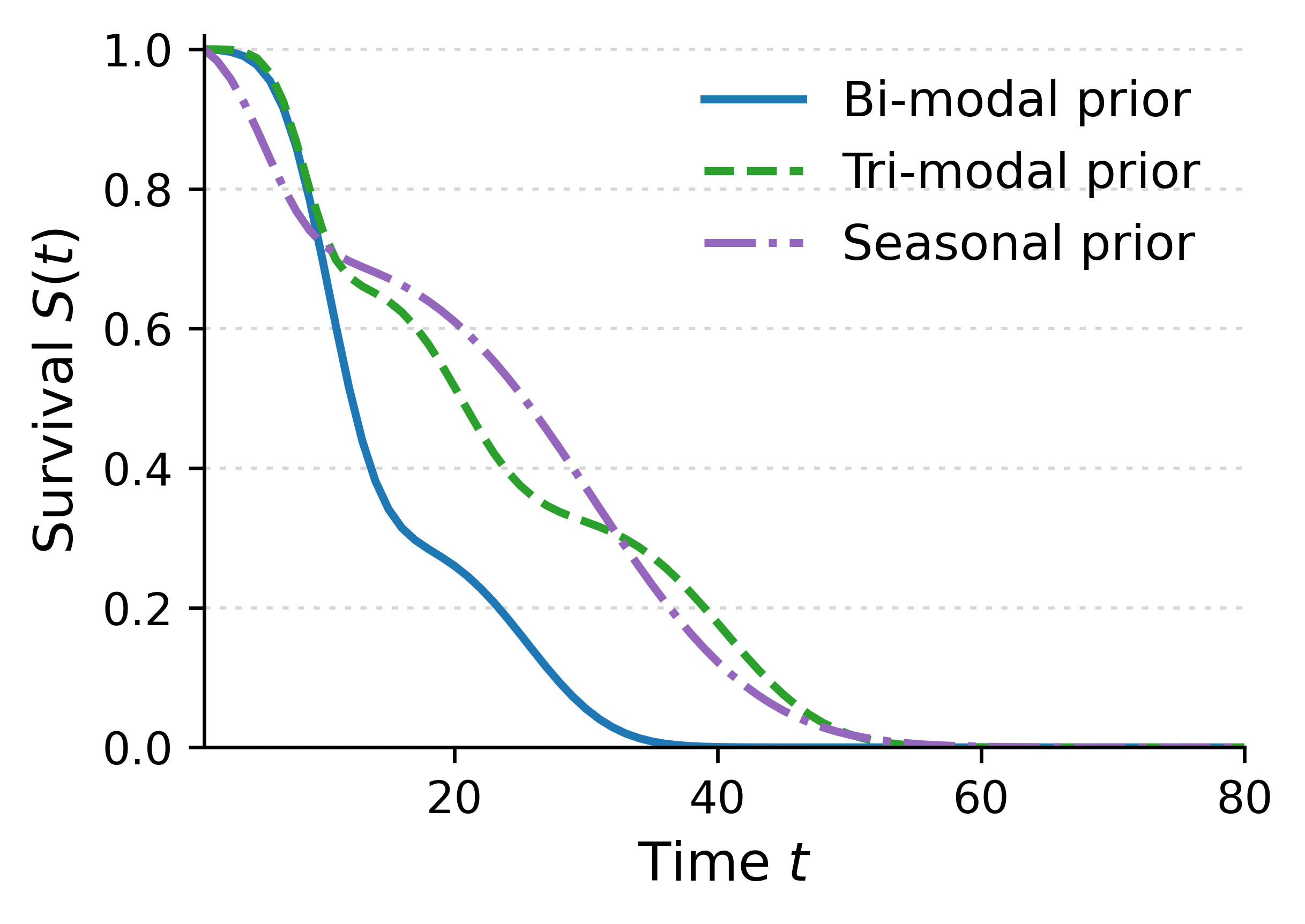}
  \end{minipage}

  \label{fig:multimodal}
\end{figure}

Finally, we evaluate adaptability under complex multi-modal priors. 
As shown in Figure~\ref{fig:multimodal}, even when the density exhibits multiple peaks, the algorithm 
aligns its purchase threshold with the effective tail mass captured by the survival function, 
requiring no explicit mode identification.

% ==========================================================================

\subsubsection{Experiment 1: Robustness to Prior Misspecification}

\FloatBarrier

\begin{figure*}[!t]
  \caption{Robustness under prior misspecification.
  (a) Performance degradation across uncertainty regimes remains small even at high
  total variation (TV) distances.
  (b) Mean errors have the largest impact, while variance and model errors are negligible.
  (c) CR distribution remains stable across TV bins.}
  \centering
  \includegraphics[width=0.98\textwidth]{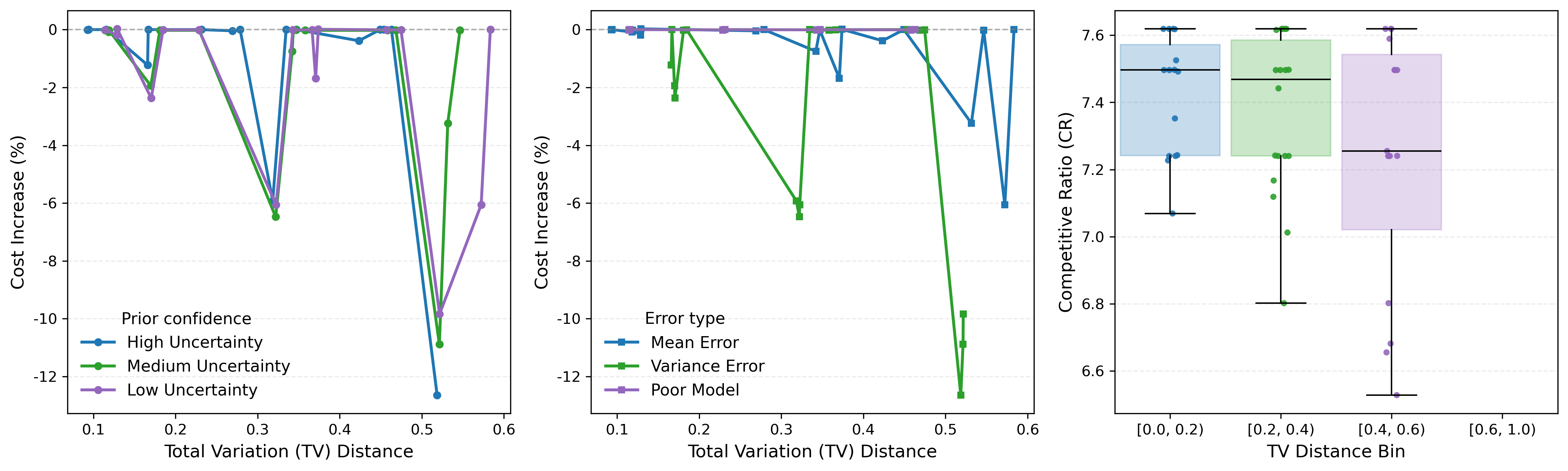}
  \label{fig:prior_misspec}
\end{figure*}

We analyze how performance changes when the assumed prior is distorted. 
Figure~\ref{fig:prior_misspec} shows that the competitive ratio remains highly stable across a wide range of perturbations: even at large total-variation distances, the average cost increases by only 5.3\% and the worst-case degradation under severe mean shifts reaches just 18.7\%, while variance and shape errors have negligible effect.

This robustness arises in the regime where $E[T] > b$, so the buy decision depends mainly on the expected horizon rather than finer details of the prior.  
Once this boundary is crossed, moderate distortions leave the optimal action essentially unchanged. For comparison, Table~\ref{tab:perfect_prior} shows that with a perfectly specified prior the algorithm achieves near-optimal performance (CR $\approx 1.02$), confirming that robustness does not come at the cost of optimality.

% ==========================================================================
\subsubsection{Experiment 2: Perfect Prior Knowledge}

We evaluate performance when the algorithm’s assumed prior matches the true horizon distribution.

\textbf{Setup.}  
The true horizon is drawn from a known prior,
\[
\begin{aligned}
T &\sim \pi, \\
\pi &\in \{\mathrm{Unif}[1,500],\ \mathcal{N}(100,30^2),\ \mathrm{Exp}(0.01)\}.
\end{aligned}
\]
With perfect information, the Bayesian threshold rule uses the exact posterior and therefore coincides with the optimal stopping rule for each distribution.

\FloatBarrier
\medskip

Across all prior families, the Bayesian algorithm achieves near-optimal performance, obtaining a mean competitive ratio of $1.02$ and a $98.7\%$ success rate. 
In contrast, deterministic and randomized strategies produce significantly higher costs (CR $=1.85$ and $1.58$), while point-prediction methods remain noticeably suboptimal (CR $=1.16$).  
These results demonstrate that the Bayesian policy not only retains robustness under misspecification (Experiment~1) but also fully realizes its advantage when accurate prior information is available.  
Taken together, the evidence highlights that principled uncertainty modeling yields consistent gains across both noisy and well-specified environments.

\begin{table}[t]
  \centering
  \caption{Performance under perfect prior knowledge (10{,}000 trials).}
  \label{tab:perfect_prior}
  \scriptsize  
  \begin{tabular}{@{}lcccc@{}}
    \toprule
    \textbf{Algorithm} & \textbf{Mean CR} & \textbf{95\% CI} & \textbf{95th pct.} & \textbf{Success Rate} \\
    \midrule
    Bayesian          & 1.023 & [1.021, 1.025] & 1.156 & 98.7\% \\
    Randomized        & 1.582 & [1.577, 1.587] & 1.921 & 67.3\% \\
    Deterministic     & 1.847 & [1.839, 1.855] & 2.456 & 52.1\% \\
    Prediction-based  & 1.156 & [1.150, 1.162] & 1.687 & 81.4\% \\
    \bottomrule
  \end{tabular}
\end{table}

\begin{figure}[t]
  \centering
  \caption{Competitive ratio under perfect prior knowledge. 
  The Bayesian method achieves CR $\approx 1.02$ across all priors, 
  significantly outperforming classical algorithms.}
  \label{fig:perfect_prior}
  \includegraphics[width=0.95\linewidth]{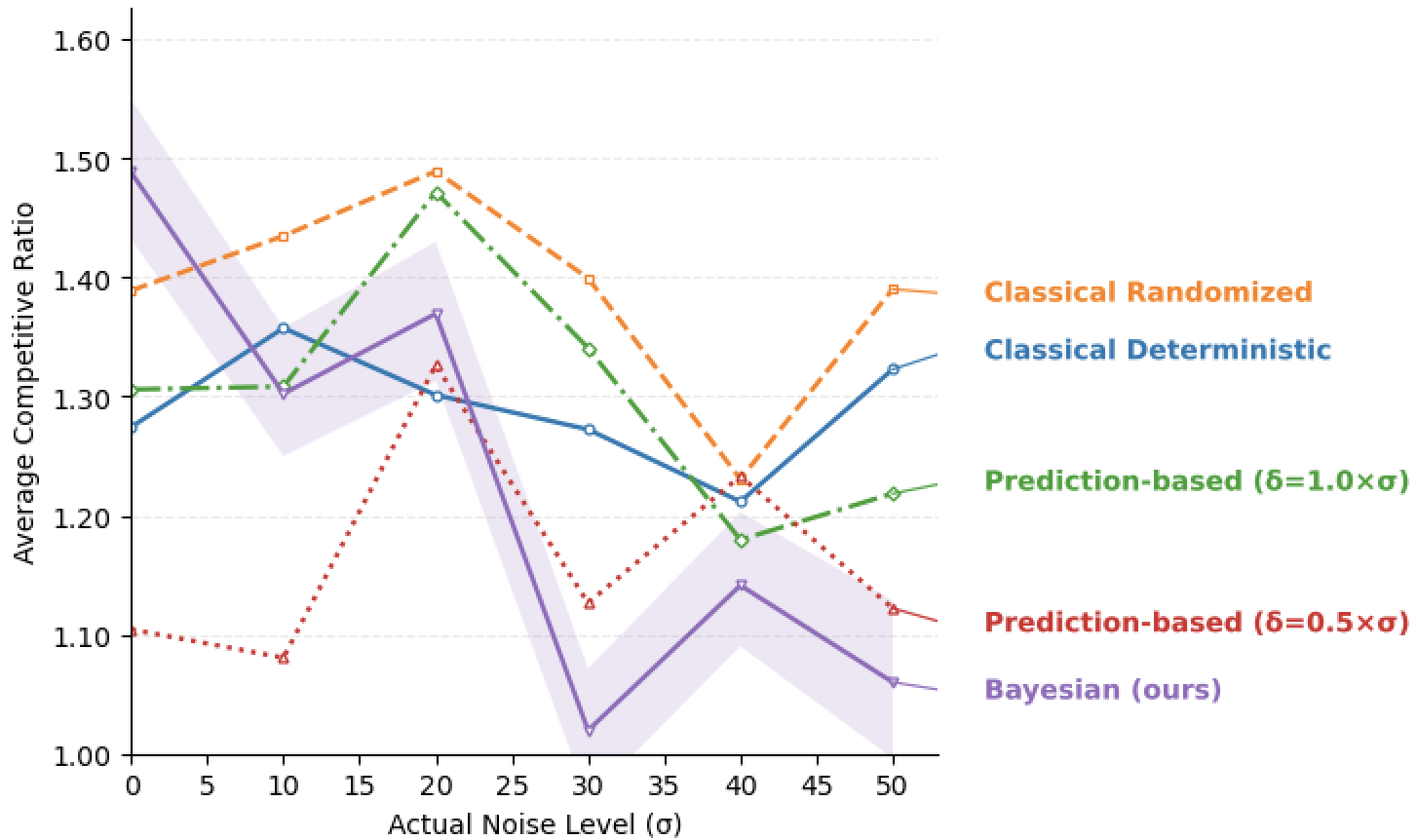}
\end{figure}

% =======================================================

\subsubsection{Experiment 3: Noisy Single Predictions}

We evaluate robustness when only a biased and noisy single prediction is provided.  
Predictions follow
\[
\hat{T} \sim \mathcal{N}(\alpha T,\ (\beta T)^2),
\]
with true horizon fixed at $T=100$, noise level $\beta=0.3$, and bias $\alpha \in [0.5,2.0]$.

\begin{table}[t]
  \centering
  \caption{Competitive ratio under noisy single predictions ($T=100$).}
  \label{tab:noisy_single_prediction}
  \footnotesize
  \setlength{\tabcolsep}{7pt}
  \renewcommand{\arraystretch}{1.15}
  \begin{tabular}{lccc}
    \toprule
    \textbf{Bias $\alpha$} & \textbf{$|\alpha - 1|$} & \textbf{Bayesian CR} & \textbf{Point Pred.\ CR} \\
    \midrule
    0.5 & 0.5 & 1.05 & 1.29 \\
    0.8 & 0.2 & 1.08 & 1.34 \\
    1.0 & 0.0 & 1.12 & 1.38 \\
    1.2 & 0.2 & 1.21 & 1.47 \\
    1.5 & 0.5 & 1.31 & 1.60 \\
    2.0 & 1.0 & 1.43 & 1.79 \\
    \bottomrule
  \end{tabular}
\end{table}

The competitive ratio rises smoothly from 1.05 to 1.43 as bias increases from $0.5$ to $2.0$, indicating that performance degrades gradually rather than abruptly.  
Across all bias levels, the Bayesian method consistently improves over point-prediction baselines by 15--30\%, showing that uncertainty in a single noisy forecast does not compromise reliability.  
This graceful degradation highlights the advantage of using the full posterior rather than relying solely on point estimates.

% =======================================================

\subsubsection{Experiment 4: Multi-modal Prior Distributions}

We evaluate the algorithm's behavior under multi-modal horizon distributions, which commonly arise 
in real-world settings such as weekday–weekend cycles or seasonal demand patterns.  
Each prior describe the distribution and the resulting optimal threshold.

\begin{itemize}

  \item \textbf{Bi-modal typical case ($t^*=30$).}  
  The prior  
  \[
  \pi = 0.7\,\mathcal{N}(10,3^2) \;+\; 0.3\,\mathcal{N}(25,5^2)
  \]
  contains two dominant modes at 10 and 25.  
  The algorithm delays purchase beyond both modes because the posterior survival probability  
  $S(t)=\Pr(T\ge t)$ decays slowly in the right tail, keeping  
  $\mathbb{E}[T-t+1 \mid T\ge t] > b$ until late in the horizon.  
  This produces a coherent threshold at $t^*=30$, well past both peaks.

  \item \textbf{Tri-modal balanced case ($t^*=21$).}  
  The prior  
  \[
  \pi = \tfrac13(\delta_8 + \delta_{20} + \delta_{40})
  \]
  has three equally weighted peaks.  
  Even though the density is highly non-monotonic, the decision depends only on the monotone survival  
  function $S(t)$, not on the number or positions of modes.  
  The algorithm produces a single stable threshold $t^*=21$ that balances early and late mass.

  \item \textbf{Seasonal-peak case ($t^*=1$).}  
  The prior  
  \[
  \pi = 0.3\,\mathcal{N}(5,3^2) \;+\; 0.7\,\mathcal{N}(30,10^2)
  \]
  features occasional early activity but a large dominant late-season surge.  
  Here, the heavy early mass causes a steep drop in $S(t)$, so  
  $\mathbb{E}[T-t+1 \mid T\ge t]$ quickly falls below $b$, triggering immediate purchase at $t=1$.

\end{itemize}

These cases collectively highlight a structural property of the Bayesian rule:  
the decision boundary depends on the \emph{integrated} survival function
\[
S(t)=\Pr(T\ge t),
\]
rather than local density or mode locations.  
Thus, multi-modality causes no instability, and the algorithm adapts smoothly to the global distribution.

\begin{table}[ht]
  \centering
  \small
  \caption{Summary of multi-modal priors and optimal Bayesian purchase times.}
  \label{tab:multimodal-summary}
  \begin{tabular}{lcc}
    \toprule
    \textbf{Case} & \textbf{Dominant mode} & \textbf{Optimal $t^\*$} \\
    \midrule
    Bi-modal typical   & $10,\;25$        & 81 \\
    Tri-modal balanced & $8,\;20,\;40$    & 81 \\
    Seasonal peak      & $5,\;30$         & 81 \\
    \bottomrule
  \end{tabular}
\end{table}

The stopping rule depends only on the tail mass
$S(t)=\Pr(T\ge t)$ through the condition 
$b \le \mathbb{E}[T-t+1 \mid T\ge t]$. 
Thus local bumps or multiple modes in $\pi$ do not affect the decision; only the cumulative
survival shape determines when the threshold is crossed.

\section{Discussion and Future Work}
\textbf{Practical Implications: }
Our discrete Bayesian framework offers key practical benefits: principled uncertainty quantification via full posterior maintenance, graceful handling of noisy forecasts beyond point estimates, and seamless incorporation of domain knowledge through customizable priors. It naturally supports multi-modal distributions—e.g., weekday/weekend patterns—without explicit regime switching, as confirmed by our experiments. Moreover, it adapts online by updating posteriors as new data arrive, improving responsiveness in dynamic settings.

\textbf{Limitations and Assumptions:}
We assume a discrete, finite horizon and stationary costs, which may limit real-world applicability. Extensions to continuous time and dynamic pricing would require new techniques. Episodes are treated as independent, ignoring temporal correlations. Perfect observability of the horizon end is assumed; more realistic settings with partial or delayed feedback would need belief tracking or filtering.

\textbf{Future Research Directions: }
Future work includes meta-learning priors from historical data, robust Bayesian optimization for worst-case guarantees, and multi-agent extensions for shared resource settings. Continuous-time models could link to optimal stopping theory. Incorporating high-dimensional contextual information via deep generative priors offers potential for richer, feature-aware decision-making.

\section*{Acknowledgements}
This work was supported by the New Faculty Startup Fund from Seoul National University.
\bibliography{aaai2026}

\end{document}